\documentclass[conference]{IEEEtran}
\IEEEoverridecommandlockouts
\usepackage{cite}
\usepackage{amsmath,amssymb,amsfonts}
\usepackage{algorithm,algorithmic}
\usepackage{graphicx}
\usepackage{subfigure}
\usepackage{textcomp}
\usepackage{xcolor}
\usepackage{hyperref}
\usepackage{array}
\def\BibTeX{{\rm B\kern-.05em{\sc i\kern-.025em b}\kern-.08em
    T\kern-.1667em\lower.7ex\hbox{E}\kern-.125emX}}

\usepackage{multirow}
\usepackage{makecell}
\usepackage[nolist,nohyperlinks]{acronym}
\acrodef{FL}{Federated Learning}
\acrodef{ML}{Machine Learning}
\acrodef{EWWA-FL}{Element-Wise Weights Aggregation Method for \ac{FL}}
\acrodef{FedBoosting}{Federated Boosting}
\acrodef{FedAvg}{Federated Averaging}
\acrodef{non-IID}{Non-Independent and Identically Distributed}
\acrodef{IID}{Independent and Identically Distributed}
\acrodef{FedAdp}{Federated Adaptive Weighting}
\acrodef{SGD}{Stochastic Gradient Descent}

\begin{document}

\title{An Element-Wise Weights Aggregation Method for Federated Learning}

\author{
\IEEEauthorblockN{1\textsuperscript{st} Yi Hu}
\IEEEauthorblockA{\textit{Department of Computer Science} \\
\textit{Swansea University}\\
Swansea, United Kingdom \\
845700@swansea.ac.uk}
\and
\IEEEauthorblockN{2\textsuperscript{nd} Hanchi Ren$\ast$}
\IEEEauthorblockA{\textit{Department of Computer Science} \\
\textit{Swansea University}\\
Swansea, United Kingdom \\
hanchi.ren@swansea.ac.uk}
\and
\IEEEauthorblockN{3\textsuperscript{th} Chen Hu}
\IEEEauthorblockA{\textit{Department of Computer Science} \\
\textit{Swansea University}\\
Swansea, United Kingdom \\
2100552@swansea.ac.uk}
\and
\IEEEauthorblockN{4\textsuperscript{rd} Jingjing Deng}
\IEEEauthorblockA{\textit{Department of Computer Science} \\
\textit{Durham University}\\
Durham, United Kingdom \\
jingjing.deng@durham.ac.uk}
\and
\IEEEauthorblockN{5\textsuperscript{th} Xianghua Xie$\ast$}
\IEEEauthorblockA{\textit{Department of Computer Science} \\
\textit{Swansea University}\\
Swansea, United Kingdom \\
x.xie@swansea.ac.uk}
\and
\thanks{$\ast$ The corresponding authors are Hanchi Ren and Xianghua Xie.}

}

\maketitle

\begin{abstract}
Federated learning (FL) is a powerful Machine Learning (ML) paradigm that enables distributed clients to collaboratively learn a shared global model while keeping the data on the original device, thereby preserving privacy. A central challenge in FL is the effective aggregation of local model weights from disparate and potentially unbalanced participating clients. Existing methods often treat each client indiscriminately, applying a single proportion to the entire local model. However, it is empirically advantageous for each weight to be assigned a specific proportion. This paper introduces an innovative Element-Wise Weights Aggregation Method for Federated Learning (EWWA-FL) aimed at optimizing learning performance and accelerating convergence speed. Unlike traditional FL approaches, EWWA-FL aggregates local weights to the global model at the level of individual elements, thereby allowing each participating client to make element-wise contributions to the learning process. By taking into account the unique dataset characteristics of each client, EWWA-FL enhances the robustness of the global model to different datasets while also achieving rapid convergence. The method is flexible enough to employ various weighting strategies. Through comprehensive experiments, we demonstrate the advanced capabilities of EWWA-FL, showing significant improvements in both accuracy and convergence speed across a range of backbones and benchmarks.
\end{abstract}


\section{Introduction}

\begin{figure}[bt!]
    \centering
    \includegraphics[width=0.97\linewidth]{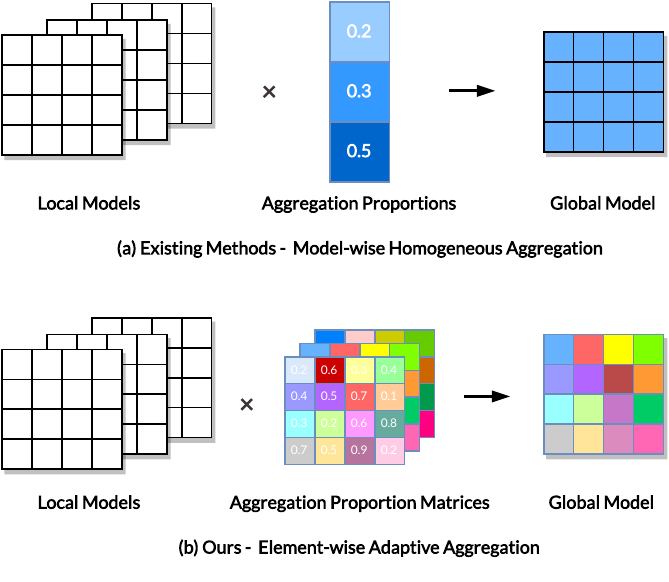}
    \caption{Illustration for our proposed EWWA-FL}
    \label{fig:intro}
\end{figure}

As the digital world continues to expand at an unprecedented rate, the world is inundated with a massive amount of data, distributed across various devices, sensors, and platforms. With the growing adoption of \ac{ML} algorithms, the demand for efficient, secure, and decentralized learning processes has become increasingly critical. \ac{FL}~\cite{mcmahan2016communication, konevcny2016federated, konevcny2016federatedlearning, mcmahan2017federated} has emerged as a promising solution to address these challenges. It enables the deployment of learning algorithms on decentralized data sources while safeguarding data privacy. \ac{FL} focuses on training \ac{ML} models across a multitude of dispersed devices or clients, each holding their own local datasets, eliminating the need for data exchange. This approach effectively addresses privacy and security concerns, as it obviates the need to transfer potentially sensitive data to a centralized location. However, a key challenge in \ac{FL} lies in the aggregation of model weights. The process of combining model weights from multiple, disparate clients is inherently complex due to the heterogeneous nature of their data distributions~\cite{nishio2019client,sery2021over,kairouz2021advances}. In an \ac{FL} network, each client utilizes its local data to train an independent model. Consequently, these local models may capture different data patterns, posing a challenge to the creation of a well-generalized global model. Various strategies, such as \ac{FedAvg}, have been developed to mitigate these issues. Nonetheless, devising an efficient and robust aggregation mechanism remains a significant challenge in the field of \ac{FL}.

There has been a surge in recent research focused on adaptive weight aggregation. Reddi \emph{et al.}~\cite{reddi2020adaptive} proposed a method called FedOpt. They provide a theoretical analysis of the model's convergence on heterogeneous data for non-convex optimization problems, as well as the relationship between dataset heterogeneity and communication efficiency. Three specific optimization methods, termed FedAdam, FedAdagrad, and FedYogi, are employed by the authors. These methods modify the global update rule of \ac{FedAvg} from one-step SGD to one-step adaptive gradient optimization. Conversely, the work presented in \cite{wang2022communication} aims to address the challenge of high communication cost in \ac{FL}. The authors propose a novel communication-efficient adaptive \ac{FL} method called FedCAMS and also provide a theoretical analysis to guarantee model convergence. They first improve upon FedAdam by incorporating AMSGrad~\cite{reddi2019convergence} with max stabilization. Both FedOpt and FedCAMS aggregate local updates and obtain an averaged gradient, upon which global model aggregation is conducted. In other words, both methods treat each client equally when generating the global updates. However, the underlying philosophy of adaptive optimization generally favors treating each individual weight independently. While FedOpt and FedCAMS both use adaptive techniques for global model aggregation, their averaging processes do not account for the varied contributions of local models trained on different datasets. This is noteworthy because different datasets result in different levels of convergence~\cite{ren2020fedboosting,wu2021fast}.

In the work of \ac{FedBoosting}~\cite{ren2020fedboosting}, the authors proposed an adaptive gradient aggregation method based on the boosting algorithm. They discovered that the generalization ability of the global model on \ac{non-IID} data is unsatisfactory due to the presence of weight divergence, particularly when employing the \ac{FedAvg} strategy. Consequently, each client participating in the training receives a unique aggregation percentage. Similarly, Wu \emph{et al.}~\cite{wu2021fast} found that in \ac{FL}, the path that minimizes the local objective does not necessarily align with the path of global minimization. This implies that each client's contribution to global aggregation will differ. Based on this observation, Wu \emph{et al.} proposed \ac{FedAdp}, a method that measures the contributions of participating clients based on the correlation between local and global gradients. All of the above studies have one thing in common: they treat all model parameters equally when aggregating the global model. Specifically, both FedOpt and FedCAMS perform a simple averaging of local model weights prior to subsequent computations. Although \ac{FedBoosting} and \ac{FedAdp} assign different proportions to each local model, they still allocate the same proportion to each parameter within these models. This approach may not be the most intuitive or efficient way to handle local models.

Key to the \ac{FL} process is the merging of model weights from different clients, which is inherently intricate and poses several challenges. The main reason for this complexity is the heterogeneity of the data distribution of the participating entities, cause each client's local dataset has different statistical properties. For example, one client's dataset may contain one or more specific classes, while another client does not. This heterogeneity can result in \ac{non-IID} data, which poses a significant challenge in aggregating local updates in a way that is representative and conducive to global model performance and generalization. To mitigate the effects of different data distributions and to ensure robust model aggregation, many sophisticated algorithms and techniques are proposed~\cite{mcmahan2016communication,reddi2020adaptive,ren2020fedboosting,wu2021fast}. Building a balanced and harmonious model requires not only rigorous mathematical or algorithmic knowledge, but also a comprehensive understanding of the differences and nuances inherent in the data landscape of different clients. Based on the findings from Wu \emph{et al.}~\cite{wu2021fast}, we would like to expand the idea to a more grained level that the elements in each local model have their personalized path to minimize the local objective. It is conceivable that each client should have a different weight, and likewise, each parameter within the local model should also have a unique weight. In the context of convex optimization for learning, the goal is to update the model's weights to achieve convergence. A model comprises various parameters, the values of which fluctuate depending on the feature space of the local dataset. Within the framework of \ac{FL}, individual local models, trained on distinct datasets, display unique patterns and directions of convergence. As a result, the same parameter across these local models may have vastly different values and may not align closely with each other. Additionally, each parameter may follow a unique trend and orientation toward convergence. As a result, using a uniform proportion to aggregate all parameters into a global model may not be the most suitable approach. Based on this understanding, we introduce \ac{EWWA-FL}, which assigns a different aggregation proportion to each parameter in the local model. Experimental results show that our method outperforms \ac{FedAvg}, FedCAMS, and FedOpt across various neural networks, benchmark datasets, and experimental settings. The contributions of this work are fourfold:

\begin{itemize}
    \item We introduce a new perspective on element-wise weight combination for \ac{FL}. This approach assigns a specific proportion to each parameter in the local model, aiming to improve aggregation. Experimental results confirm the novelty of our proposed \ac{EWWA-FL}.

    \item A comprehensive evaluation is conducted. We test the model's generalization ability using various neural networks on different benchmark datasets, employing both \ac{IID} and \ac{non-IID} strategies.

    \item The adaptive element-wise aggregation paradigm demonstrates faster convergence compared to other recent works.

    \item We disclose the implementation details of the proposed algorithm to ensure its reproducibility.

\end{itemize}

The paper is organized as follows: Section~\ref{sec:rw} reviews previous studies related to adaptive weight aggregation in \ac{FL}. Preliminaries on vanilla \ac{FL} and the Adam optimization algorithm are then discussed. Our proposed approach is elaborated upon in Section~\ref{sec:me}. Section~\ref{sec:er} provides insights into the experiments, offers in-depth discussions, and suggests potential mitigation methods. Finally, concluding remarks are presented in Section~\ref{sec:cc}.

\section{Related Work}
\label{sec:rw}

A fundamental challenge in \ac{FL} is the efficient aggregation of model weights from diverse and potentially \ac{non-IID} data sources to produce a globally consistent model. Adaptive weight aggregation addresses this challenge by assigning different proportions to local model weights based on their quality or relevance, as opposed to treating them equally. This approach recognizes the inherent heterogeneity present in real-world \ac{FL} environments. It optimizes the performance of the global model by leveraging the more informative weights from local models and potentially mitigates the negative impact of less reliable participants.

In the work~\cite{reddi2020adaptive}, the authors provide a comprehensive discussion on adaptive weight aggregation for \ac{FL} and propose a flexible framework called FedOpt. This framework is capable of incorporating multiple optimization algorithms. The authors specialize FedOpt into FedAdam, FedAdagrad, and FedYogi by employing three example optimization algorithms: Adam~\cite{kingma2014adam}, Adagrad~\cite{duchi2011adaptive}, and YOGI~\cite{zaheer2018adaptive}. This approach closely parallels the \ac{FedAvg} process, diverging only in the final stage of weight aggregation. After obtaining the averaged local gradients, denoted as $\hat{g}$, the first-order momentum matrices $m$ are computed for FedAdam, FedAdagrad, and FedYogi, as detailed in \eqref{equ:first-order}. However, the computation of the second-order variance matrices $v$ varies depending on the algorithm. Specifically, FedAdam employs \eqref{equ:fedadam}, while FedAdagrad and FedYogi utilize \eqref{equ:fedadagrad} and \eqref{equ:fedyogi}, respectively, to derive their second-order matrices.

\begin{align}
    m_r &= \beta_1 m_{r-1} + (1 - \beta_1) \hat{g}_r \label{equ:first-order} \\
    v_r &= \beta_2 v_{r-1} + (1 - \beta_2) \hat{g}_r^2 \label{equ:fedadam} \\
    v_r &= v_{r-1} + \hat{g}_r^2 \label{equ:fedadagrad} \\
    v_r &= \beta_2 v_{r-1} + (1 - \beta_2)\cdot \hat{g}_r^2 \cdot sign(v_{r-1} - \hat{g}_r^2) \label{equ:fedyogi} 
\end{align}
where, $r$ is the training round, $\beta_1$ and $\beta_2$ are two momentum parameters, $sign()$ is the symbolic functions. In the end, all those three methods employ Eq.\ref{equ:update} for weights aggregation.
\begin{align}
    \omega_r = \omega_{r-1} + \eta_r \frac{m_r}{\sqrt{v_r} + \epsilon} \label{equ:update}
\end{align}
where, $\eta_r$ is the adaptive learning rate, calculated by:
\begin{align}
    \eta_r = \eta_0 \frac{\sqrt{1 - \beta_2^{r}}}{1 - \beta_1^{r}}
\end{align}
where  $\eta_0$ denotes the initial learning rate, while $\beta_1^r$ and $\beta_2^r$ represent the $r$-th powers of the parameters $\beta_1$ and $\beta_2$, respectively. The authors provide a theoretical analysis to demonstrate the superiority of the proposed FedOpt in comparison to other methods. The primary distinction between FedOpt and our proposed method, \ac{EWWA-FL}, lies in the location of the optimization algorithm. Specifically, FedOpt employs the optimization algorithm after averaging the local models, whereas \ac{EWWA-FL} performs the optimization after each local training. As a result, FedOpt treats each client equally and assigns the same aggregation proportion to each local model through averaging. In contrast, our method treats each parameter in every local model differently. Building upon FedOpt, Wang \emph{et al.}~\cite{wang2022communication} introduced FedCAMS with the objective of reducing communication costs. The optimization algorithm in FedCAMS occupies the same position as in FedOpt, thereby ensuring that all local weights are aggregated equally.

Unlike FedOpt and FedCAMS, \ac{FedBoosting}~\cite{ren2020fedboosting} and \ac{FedAdp}~\cite{wu2021fast} assign different proportions to each local model to perform adaptive weight aggregation. \ac{FedBoosting} computes the aggregation proportion based on the results of local training $T^i_r$ and cross-validation $V^{i,j}_r$. The authors first sum all the validation results for a local model across all other local model validation datasets. Then, they calculate the weight of this sum of validation results. Finally, a \emph{Softmax} function is applied to derive the final proportion for each local model. Equations.\ref{equ:fedboosting-1}, \ref{equ:fedboosting-2}, and \ref{equ:fedboosting-3} provide the local weight aggregation proportion $p^i_r$ for the $i$-th local model in training round $r$:
\begin{align}
p^{(i)}_r &= softmax(softmax(T^{(i)}_r) \cdot \sum_{j \neq i}^N V^{(i,j)}_r) \label{equ:fedboosting-1} \\
s&oftmax(T^{(i)}_r) = \frac{exp(T^{(i)}_r)}{\sum_{j=1}^N {exp(T^{(j)}_r)}} \label{equ:fedboosting-2}\\
V^{(i,j)}_r &=
 \begin{pmatrix}
  V^{(1,1)}_r & V^{(1,2)}_r & \cdots & V^{(1,j)}_r \\
  V^{(2,1)}_r & V^{(2,2)}_r & \cdots & V^{(2,j)}_r \\
  \vdots & \vdots & \ddots & \vdots \\
  V^{(i,1)}_r & V^{(i,2)}_r & \cdots & V^{(i,j)}_r
 \end{pmatrix} \label{equ:fedboosting-3}
\end{align}

On the other hand, \ac{FedAdp} focuses on the angle of convergence between the updated local weight and the global weight. In particular, they quantify the contribution of each client in each round of global observations according to the angle $\theta^i$:
\begin{align}
    \theta^{(i)} = arccos(\frac{<G, g^{(i)}>}{||G||\cdot||g^{(i)}||})
\end{align}
where $G$ is the global gradient, $<\cdot>$ is the inner product operation and $||\cdot||$ denotes the L2 normalization. To suppress instability caused by instantaneous angular randomness, the angle $\theta^i_r$ is then averaged over previous training rounds $r$:
\begin{align}
    \hat{\theta}^{(i)}_r = \left\{\begin{array}{ccl}
                                \theta^{(i)}_r & \mbox{if} & r=1 \nonumber\\
                                \frac{r-1}{r}\hat{\theta}^{(i)}_{r-1} + \frac{1}{r}\theta^{(i)}_r & \mbox{if} & r>1
                            \end{array}\right. \nonumber
\end{align}
The authors then designed a non-linear mapping function that quantifies each client's contribution based on angular information. Inspired by the \emph{Sigmoid} function, they use a variant of \emph{Gompertz} function~\cite{gibbs2000variational}:
\begin{align}
\mathcal{F}(\hat{\theta}) = \alpha(1 - \frac{1}{exp(exp(\alpha(1 - \hat{\theta})))})
\end{align}
where $\alpha$ is a hyper-parameter. The final proportions for each local model are calculated by giving each client's contribution value into the \emph{Softmax} function.

In comparison to FedOpt and FedCAMS, although \ac{FedBoosting} and \ac{FedAdp} provide different aggregation proportions for local clients, their aggregation proportions are still at the model level. In contrast, our proposed \ac{EWWA-FL} makes progress in this regard by providing a more fine-grained, element-wise aggregation level. This feature enhances the adaptability and convergence of the global model, especially considering that the local models come from different datasets.

\section{Methodology}
\label{sec:me}

In this section, we first present a preliminary discussion on \ac{FedAvg}, as it is the most commonly used method in \ac{FL} applications. Subsequently, we briefly explain the Adam optimization algorithm, highlighting that Adam provides an adaptive learning rate for each parameter, in contrast to the \ac{SGD} algorithm. Finally, we introduce the \ac{EWWA-FL} algorithm, which enables element-wise global weight aggregation in \ac{FL}.

\subsection{FedAvg}

It is the most basic method behind all the recent proposed \ac{FL} methods. Assuming we have many clients $\mathcal{C}$ and their local datasets $\mathcal{D}$. The task is formulated as $\mathcal{F}$ with weights $\omega$. So the local gradient $g$ is:
\begin{align}
    g_i = \frac{1}{||d_i||}\nabla_\omega \sum_j\mathcal{L}(\mathcal{F}(x^{(j)}; \omega), y^{(j)});\forall i \in ||\mathcal{C}||
\end{align}
Where $||\cdot||$ denotes the L2 normalization of a vector, $x$ and $y$ are the samples and their relevant labels in the $i$-th local dataset $d_i$. The server gathers all the local gradients and conduct the averaging process to generate the next round of global model $\omega_{r}$. We assume that $||d_i|| = ||d_k||; \forall d_i,d_k \in \mathcal{D}$.
\begin{align}
    \omega_{r}=\omega_{r-1} - \frac {1}{\mathcal{C}} \sum_{i=1}^{\mathcal{C}} g_i
\end{align}

The \ac{FedAvg} algorithm~\cite{mcmahan2017communication} aims to create a unified model by averaging gradients from various local clients. While this method is effective for centralizing distributed learning, it is not without shortcomings. Specifically, inherent differences in data distributions among clients lead to diverse convergence directions for local model weights. This diversity arises from the incoherent feature spaces of the data, posing challenges for \ac{FedAvg}. When local datasets differ significantly in their data distributions, this can induce a considerable bias in local model weights. Consequently, the simplistic averaging mechanism employed by \ac{FedAvg} may not yield optimal results, particularly in the presence of significant data biases or extreme outliers. Recognizing these limitations, we have started exploring more nuanced aggregation strategies, such as weighted averages or other adaptive mechanisms. These approaches gauge the contribution of each local model based on factors like data distribution or its relevance to the overall learning objective~\cite{zhao2018federated,li2019convergence,xu2022acceleration}. Employing these refined techniques produces a global model that is both resilient and accurate, skillfully navigating the complexities of differing data distributions and overcoming some limitations inherent to traditional \ac{FedAvg} methods.

\subsection{Element-Wise Aggregation for FL}

Prior to detail our proposed method, we would like to briefly introduce the Adam algorithm~\cite{kingma2014adam} firstly. It is a \ac{SGD} optimization method based on the momentum idea. Before each iteration, the first-order and second-order moments of the gradient are computed and the sliding average is computed to update the current parameters. This idea combines the ability of the Adagrad~\cite{duchi2011adaptive} algorithm to handle sparse data with the properties of the RMSProp~\cite{hinton2012neural} algorithm to deal with non-smooth data. Finally, it achieves very good test performance on both traditional convex optimization problems and deep learning optimization problems. More details of Adam is shown in Algorithm~\ref{alg:adam}.

\begin{algorithm}[ht!]
    \caption{Adam}
    \begin{algorithmic}[1]
        \REQUIRE initial learning rate $\alpha$
        \REQUIRE exponential decay rates $\beta_1, \beta_2 \in [0,1]$
        \REQUIRE maximum iteration number $I$
        \STATE initial weights $\omega_0$
        \STATE initial $1^{st}$-order moment vector $m_0 \leftarrow 0$
        \STATE initial $2^{nd}$-order moment vector $v_0 \leftarrow 0$
        \FOR{each iteration $i = 1, 2, 3, ..., I$}
            \STATE $g_i \leftarrow \nabla_\omega\mathcal{L}(\mathcal{F}(\omega_{i-1}))$ $\hfill\blacktriangleright$ get gradients at iteration $i$
            \STATE $m_i \leftarrow \beta_1\cdot m_{i-1} + (1 - \beta_1)\cdot g_t$ $\hfill\blacktriangleright$ update biased $1^{st}$-order moment estimate
            \STATE $v_i \leftarrow \beta_2\cdot m_{i-1} + (1 - \beta_2)\cdot (g_t \odot g_t)$ $\hfill\blacktriangleright$ update biased $2^{nd}$-order moment estimate
            \STATE $\hat{m_i} \leftarrow \frac{m_i}{1 - \beta_1^i}$ $\hfill\blacktriangleright$ compute biased-corrected $1^{st}$-order moment estimate 
            \STATE $\hat{v_i} \leftarrow \frac{v_i}{1 - \beta_2^i}$ $\hfill\blacktriangleright$ compute biased-corrected $2^{nd}$-order moment estimate
            \STATE $\omega_i \leftarrow \omega_{i-1} - \frac{\alpha}{\sqrt{\hat{v_i}} + \epsilon} \cdot \hat{m_i}$ $\hfill\blacktriangleright$ update weights
        \ENDFOR
        \RETURN $\omega_I$
    \end{algorithmic}
 \label{alg:adam}
\end{algorithm}

\begin{figure*}[bt!]
    \centering
    \includegraphics[width=0.90\linewidth]{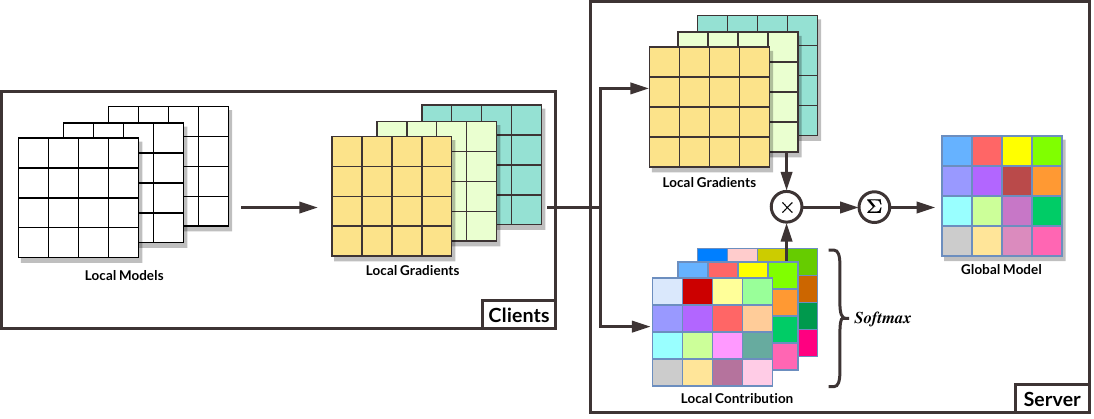}
    \caption{Diagram of \ac{EWWA-FL}. The left part is the client that performs the local training. The right side is the server calculating the local contributions and aggregating a new global model based on the local contributions and gradients. All calculations on the server end are done on an element-by-element basis.}
    \label{fig:g2}
\end{figure*}

All of the above works~\cite{mcmahan2017communication,reddi2020adaptive,wang2022communication,ren2020fedboosting,wu2021fast} either average the local models or assign dynamic proportions to the entire local model for global model aggregation. The learning process for deep learning model can be viewed as a convex optimization problem where the weights in the model are trained to reach a minimum point. The basic components of the model are a number of parameters whose values can vary greatly with the different feature spaces of different local datasets. In the \ac{FL} scenario, local models trained on different local datasets obtain different degrees and directions of convergence. In these local models, the values of the same parameters may be completely different or even not close to each other, leading to convergence heterogeneity in the local models. Therefore, giving one proportion for all the parameters in the local model is not the best way to aggregate the global model. From another perspective, in contrast to \ac{SGD}, the adaptive learning methods calculate the learning rate from the point of view of the elements. That is to say, in common model training scenarios, Adam provides an element-wise adaptive learning rate. We would like to follow Adam's idea and introduce element-level adaptive aggregation to \ac{FL}. As shown in Figure~\ref{fig:g2}, the left part is the client that performs the local training. The right side is the server, which calculates the local contributions and aggregates a new global model based on the local contributions and gradients. All computations on the server end are performed on an element-by-element basis. Specifically, Algorithm~\ref{alg:ewwa-fl} is the pseudo-code of our proposed \ac{EWWA-FL}. In the first round of global training, the global weights $\omega_0$, first-order moment vectors $m_0$ and second-order moment vectors $v_0$ are initialized. Then, the server end assigns weights to each local model for local training. Once the local training is complete, The server collects the gradients of all local models to update the biased first-order moments and second-order moments $m_r$ and $v_r$ and compute the unbiased estimates $\hat{m}_r$ and $\hat{v}_r$. Then obtain the contribution $b_r$ of the local model to the new global model. Finally, the aggregation proportion for each local model is computed using \emph{Softmax}. All the computations related to the first-order moment, second-order moment, local contribution and final aggregation proportion are element-wise, which means each parameter in each local model will receive a specific aggregation proportion in each round, rather than one proportion for all parameters in the local model.

\begin{algorithm}[ht!]
    \caption{EWWA-FL}
    \begin{algorithmic}[1]
        \REQUIRE initial local learning rate $\alpha$
        \REQUIRE exponential decay rates $\beta_1, \beta_2 \in [0,1]$
        \REQUIRE global training round $R$
        \STATE initial weights $\omega_0$
        \STATE initial $1^{st}$-order moment vector $m_0 \leftarrow 0$
        \STATE initial $2^{nd}$-order moment vector $v_0 \leftarrow 0$
        \FOR{each round $r = 1, 2, ..., R$}
            \FOR{each client $c \in C$}
                \STATE $g_r^{(c)} \leftarrow \nabla_{\omega}\mathcal{L}(\mathcal{F}(\omega_{r-1}))$
                \STATE $m_r^{(c)} \leftarrow \beta_1\cdot m_{r-1} + (1 - \beta_1)\cdot g_r^{(c)}$
                \STATE $v_r^{(c)} \leftarrow \beta_2\cdot m_{r-1} + (1 - \beta_2)\cdot (g_r^{(c)}\odot g_r^{(c)})$
                \STATE $\hat{m}_r^{(c)} \leftarrow \frac{m_r^{(c)}}{1 - \beta_1^r}$
                \STATE $\hat{v}_r^{(c)} \leftarrow \frac{v_r^{(c)}}{1 - \beta_2^r}$
                \STATE $b_r^{(c)} \leftarrow \frac{\alpha}{\sqrt{\hat{v_r^{(c)}}} + \epsilon}\cdot \hat{m}_r^{(c)}$
            \ENDFOR
        \STATE $p_r^{(c)} \leftarrow \frac{exp(b_r^{(c)})}{\sum_{i=c}^{C}exp(b_r^{(i)})}; \forall c \in C$
        \STATE $G_r \leftarrow \sum_{i=c}^C p_r^{(i)}\cdot g_r^{(i)}$
        \ENDFOR
    \end{algorithmic}
 \label{alg:ewwa-fl}
\end{algorithm}

\section{Experiments}
\label{sec:er}

In this section, we first describe the settings of all experiments. Then, we introduce the backbone neural networks and the datasets for benchmark evaluation. After that, we present multiple sets of experiments to access the performance of our proposed \ac{EWWA-FL} against other state-of-the-art methods.


\begin{table*}[!ht]
\setlength{\tabcolsep}{8pt}
\begin{center}
\caption{Top-1 classification accuracy (\%) across different methods, backbone neural networks and benchmark datasets with \ac{IID} distribution on local clients. ``C-10'', ``C-100'' and ``IC-12'' stand for CIFAR-10, CIFAR-100 and ILSVRC2012, respectively. FedAMS is derived from FedCAMS and has no efficient communication settings. The red ones are the highest accuracy and the blue ones are the next highest. Crossed symbols indicate experimental failure, \emph{i.e.} no global model converged in five trials.}
\label{tab:accuracy}
\begin{tabular}{c c|c|c c|c c|c c c|c c c}
\hline
\multicolumn{2}{c|}{\textbf{Model}} & \makecell[c]{\emph{LeNet}\\(32*32)} & \multicolumn{2}{c|}{\makecell[c]{\emph{ResNet-20}\\(32*32)}} & \multicolumn{2}{c|}{\makecell[c]{\emph{ResNet-32}\\(32*32)}} & \multicolumn{3}{c|}{\makecell[c]{\emph{ResNet-18}\\(224*224)}} & \multicolumn{3}{c}{\makecell[c]{\emph{ResNet-34}\\(224*224)}} \\
\hline
\multicolumn{2}{c|}{\textbf{Dataset}} & MNIST & C-10 & C-100 & C-10 & C-100 & C-10 & C-100 & IC-12 & C-10 & C-100 & IC-12 \\
\hline
\multicolumn{2}{c|}{\textbf{FedAvg}}& \textcolor{blue}{98.14} & \textcolor{red}{91.20} & 58.58 & \textcolor{red}{91.37} & 61.91 & 89.50 & 68.59 & 65.50 & 89.27 & 68.30 & 67.91 \\
\multicolumn{2}{c|}{\textbf{FedAMS}} & \textcolor{red}{98.77} & 86.57 & 54.58 & 87.21 & 54.23 & \textcolor{red}{91.59} & 66.07 & $\times$ & \textcolor{red}{91.91} & 65.41 & $\times$ \\
\multicolumn{2}{c|}{\textbf{FedCAMS}} & $\times$ & 76.77 & 41.65 & 84.7 & 41.77 & \textcolor{blue}{91.35} & 66.62 & $\times$ & \textcolor{blue}{91.44} & 66.41 & $\times$ \\
\hline
\multirow{3}{*}{\textbf{FedOpt}} & \textbf{Adam} & $\times$ & 73.59 & 22.31 & 74.84 & $\times$ & 78.07 & $\times$ & $\times$ & 81.71 & $\times$ & $\times$ \\
& \textbf{Adagrad} & $\times$ & 64.63 & $\times$ & 67.36 & $\times$ & $\times$ & $\times$ & $\times$ & $\times$ & $\times$ & $\times$ \\
& \textbf{YOGI} & $\times$ & 65.93 & $\times$ & 71.90 & $\times$ & 73.88 & $\times$ & $\times$ & 71.8 & $\times$ & $\times$ \\
\hline
\multirow{3}{*}{\textbf{EWWA-FL}} & \textbf{Adam} & 98.00 & 89.73 & \textcolor{blue}{64.14} & \textcolor{blue}{90.17} & \textcolor{blue}{65.63} & 90.77 & \textcolor{red}{70.98} & \textcolor{red}{65.64} & 91.23 & \textcolor{blue}{70.15} & 68.23 \\
& \textbf{Adagrad} & 97.99 & \textcolor{blue}{89.74} & \textcolor{red}{64.16} & \textcolor{blue}{90.17} & \textcolor{red}{65.84} & 91.17 & 70.34 & \textcolor{red}{65.64} & 91.13 & 69.77 & \textcolor{red}{68.33} \\
& \textbf{YOGI} & 98.00 & 89.43 & 64.10 & 90.10 & 65.38 & 90.88 & \textcolor{blue}{70.78} & \textcolor{blue}{65.54} & 91.13 & \textcolor{red}{70.34} & \textcolor{blue}{68.27} \\
\hline
\end{tabular}
\end{center}
\end{table*}

\begin{table*}[!ht]
\setlength{\tabcolsep}{5pt}
\begin{center}
\caption{Percentage (\%) of top-1 classification accuracy across methods, backbone neural networks, and benchmark datasets in \ac{non-IID} conditions on local clients. The values highlighted in red demonstrate the highest performance. The global aggregation optimization algorithm for FedOpt and \ac{EWWA-FL} is Adam.}
\label{tab:noniid-accuracy}
\begin{tabular}{c|c|c c|c c|c c c|c c c}
\hline
\textbf{Model} & \makecell[c]{\emph{LeNet}\\(32*32)} & \multicolumn{2}{c|}{\makecell[c]{\emph{ResNet-20}\\(32*32)}} & \multicolumn{2}{c|}{\makecell[c]{\emph{ResNet-32}\\(32*32)}} & \multicolumn{3}{c|}{\makecell[c]{\emph{ResNet-18}\\(224*224)}} & \multicolumn{3}{c}{\makecell[c]{\emph{ResNet-34}\\(224*224)}} \\
\hline
\textbf{Dataset} & MNIST & C-10 & C-100 & C-10 & C-100 & C-10 & C-100 & IC-12 & C-10 & C-100 & IC-12 \\
\hline
\textbf{FedAvg}& 96.30 & 72.57 & 55.49 & 75.76 & 57.11 & 82.55 & 66.26 & 50.68 & 82.64 & 65.02 & 46.13 \\
\textbf{FedAMS}& 96.78 & 54.78 & 37.33 & 54.46 & 39.71 & 55.71 & 33.71 & 29.26 & 44.13 & 32.12 & 24.09 \\
\textbf{FedCAMS}& $\times$ & 47.76 & $\times$ & 46.47 & 26.87 & 40.03 & 25.57 & $\times$ & 41.27 & 24.27 & $\times$ \\
\textbf{FedOpt}& $\times$ & 61.16 & $\times$ & 46.75 & $\times$ & $\times$ & $\times$ & $\times$ & $\times$ & $\times$ & $\times$ \\
\hline
\textbf{EWWA-FL}& \textcolor{red}{96.88} & \textcolor{red}{76.04} & \textcolor{red}{56.07} & \textcolor{red}{78.50} & \textcolor{red}{57.56} & \textcolor{red}{83.86} & \textcolor{red}{66.74} & \textcolor{red}{51.67} & \textcolor{red}{84.36} & \textcolor{red}{65.97} & \textcolor{red}{52.50} \\
\hline
\end{tabular}
\end{center}
\end{table*}

\begin{figure*}[ht!]
\centering
    \subfigure[ResNet-18, CIFAR-10, IID]{
    \centering
    \includegraphics[width=0.43\linewidth]{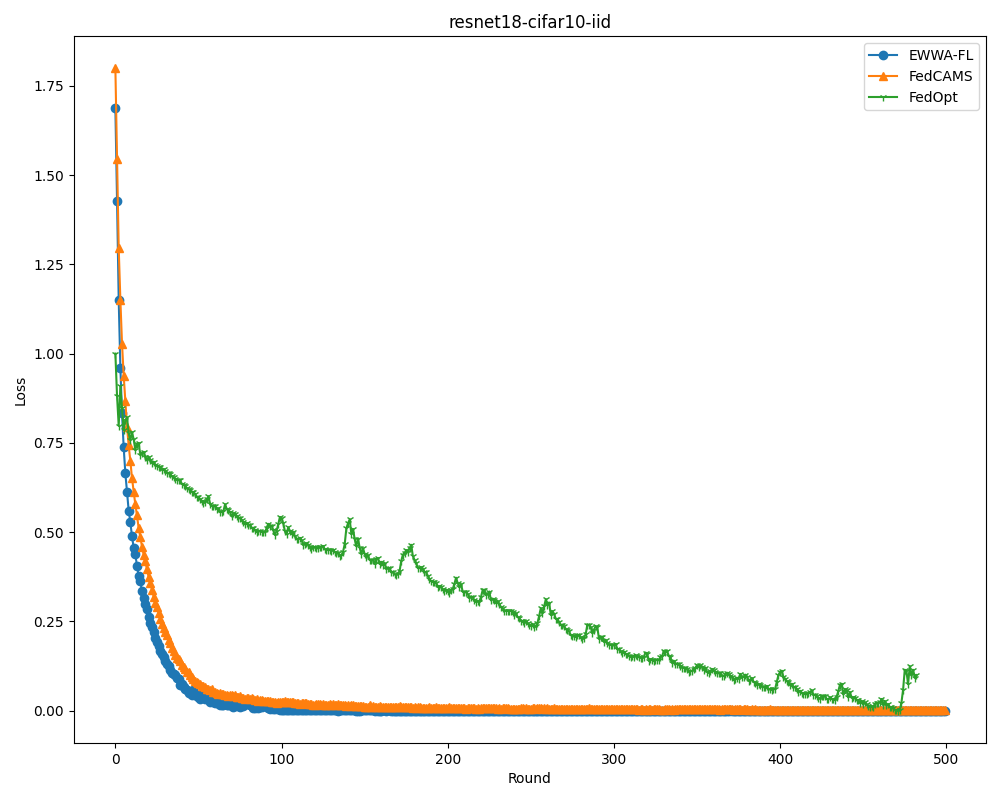}
    }
    \subfigure[ResNet-18, CIFAR-10, non-IID]{
    \centering
    \includegraphics[width=0.43\linewidth]{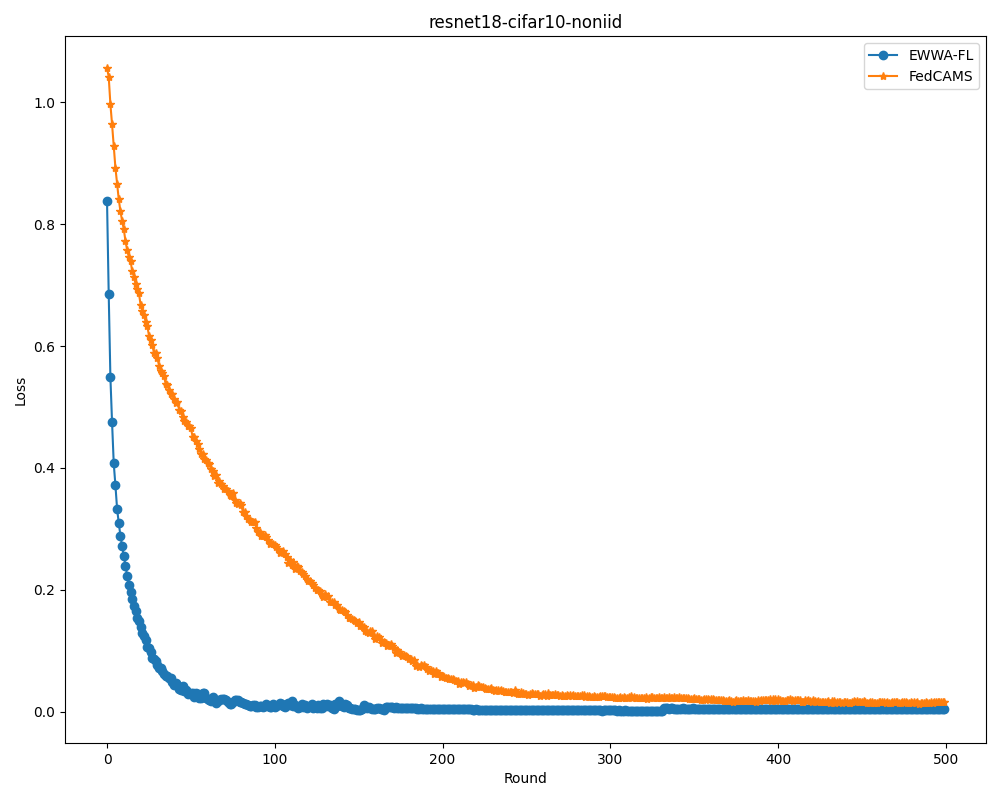}
    }
    \subfigure[ResNet-18, CIFAR-100, IID]{
    \centering
    \includegraphics[width=0.43\linewidth]{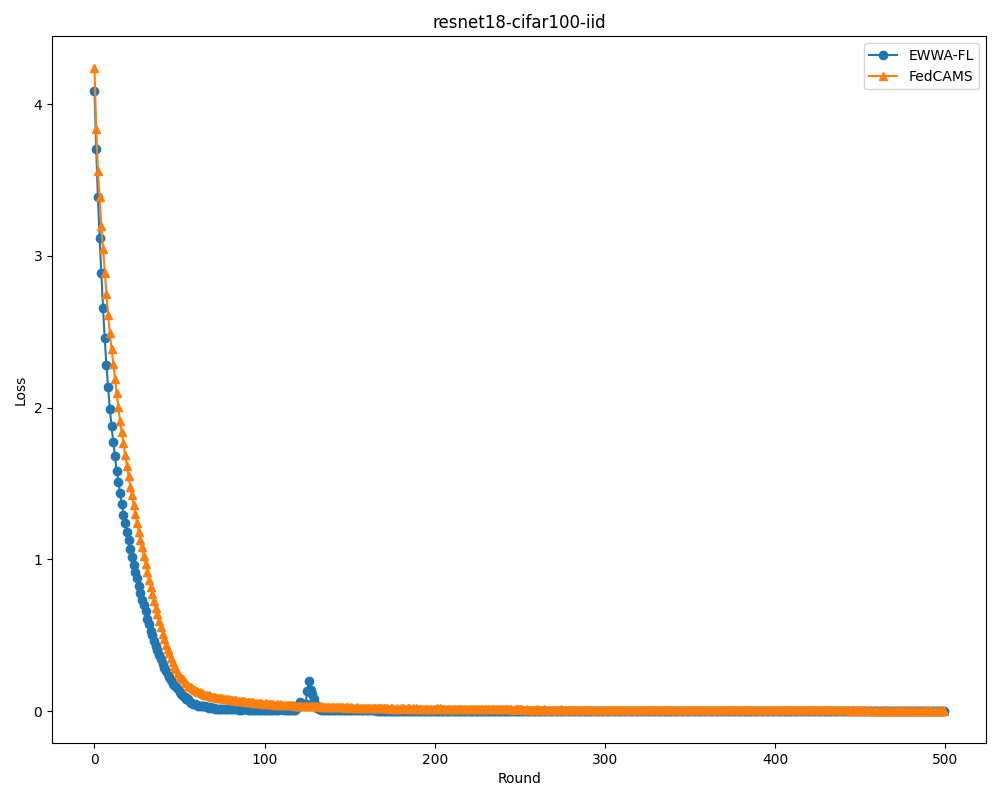}
    }
    \subfigure[ResNet-18, CIFAR-100, non-IID]{
    \centering
    \includegraphics[width=0.43\linewidth]{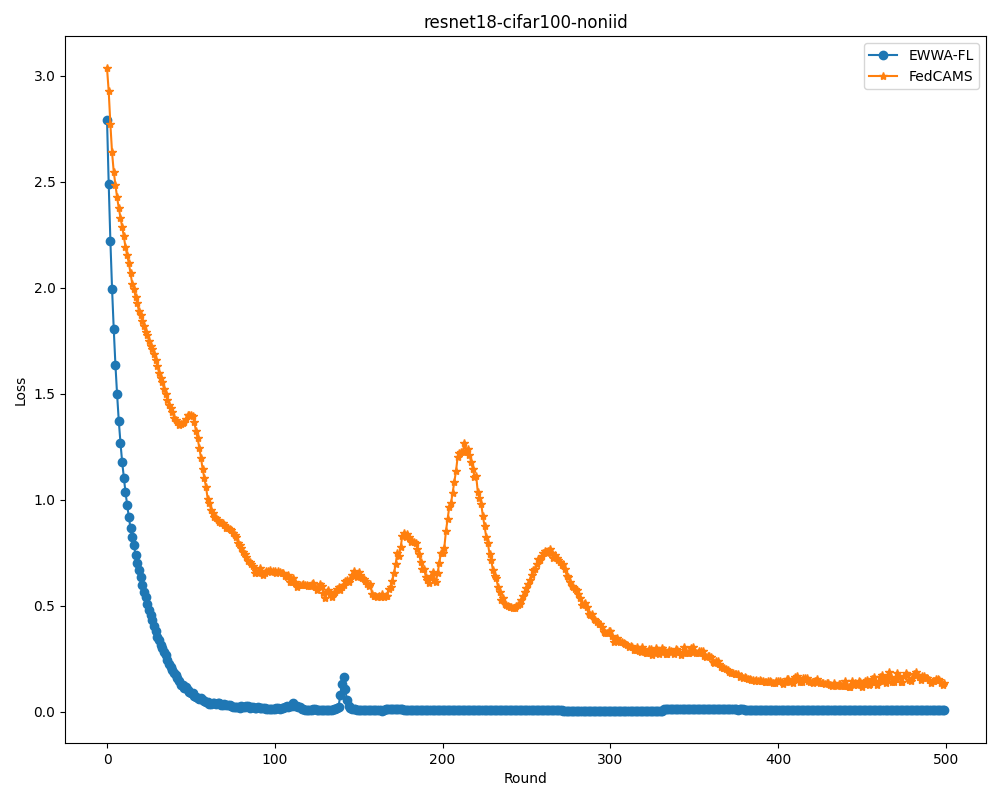}
    }
    \subfigure[ResNet-32, CIFAR-10, IID]{
    \centering
    \includegraphics[width=0.43\linewidth]{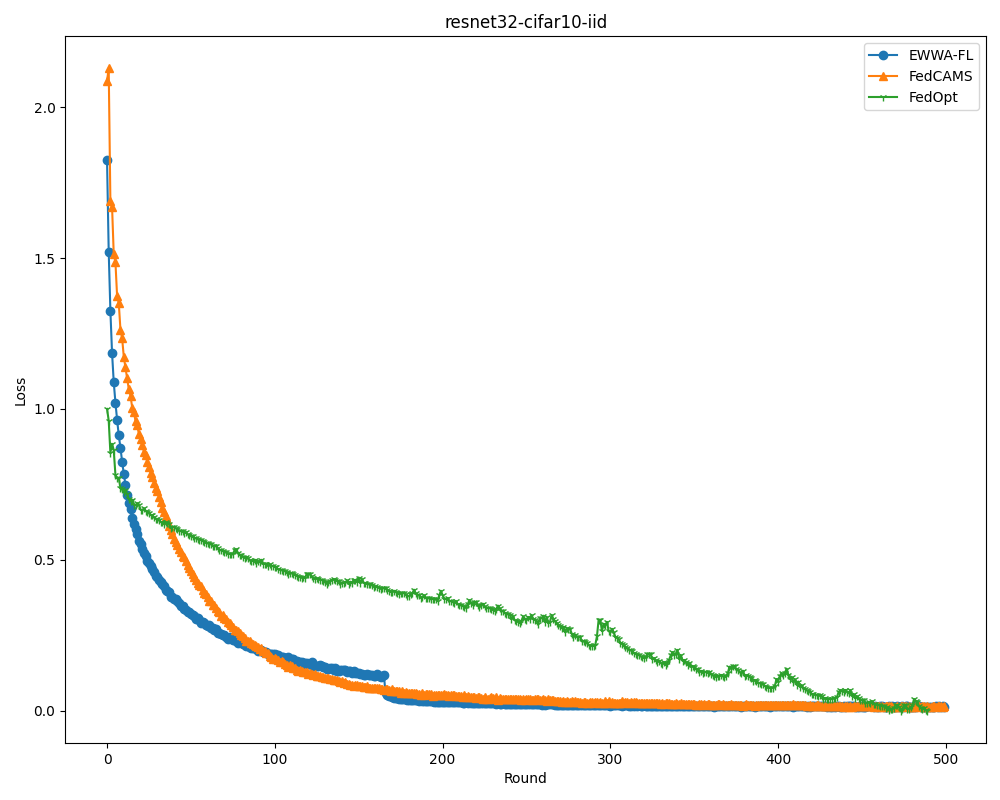}
    }
    \subfigure[ResNet-32, CIFAR-10, non-IID]{
    \centering
    \includegraphics[width=0.43\linewidth]{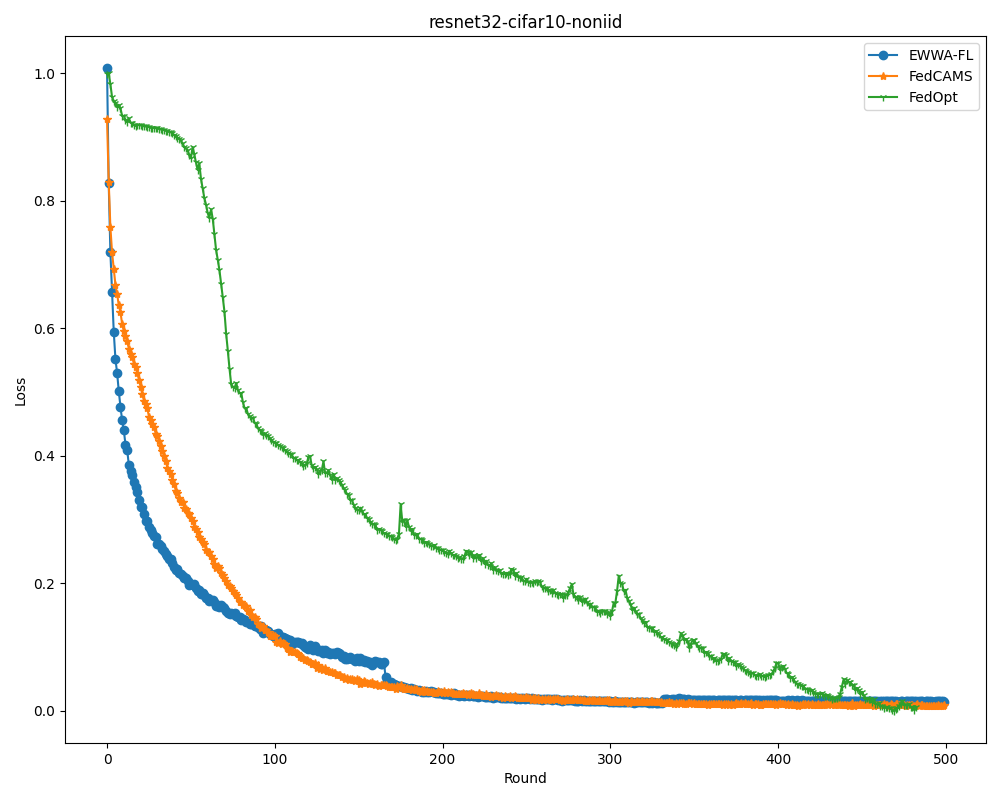}
    }
\caption{Visualization of the average training loss across all local clients is shown. The horizontal axis represents the number of global aggregation rounds, while the vertical axis indicates the average loss for the current round. The blue line represents our proposed \ac{EWWA-FL}, the orange line is from FedCAMS, and the green line represents FedOpt.}
\label{fig:cs}
\end{figure*}

\subsection{Settings, Backbones and Datasets}

We utilized the PyTorch framework~\cite{paszke2019pytorch} to implement all neural network models. For anyone interested in replicating our results, the source code is open to the public and can be accessed here\footnote{\url{https://github.com/Rand2AI/EWWA-FL}}. For our proposed \ac{EWWA-FL} method, the global aggregation learning rate for Adam, Adagrad, and Yogi was set at $1.0$ based on the FedOpt reference~\cite{reddi2020adaptive}. We chose the two momentum parameters of $0.9$ and $0.999$. Local training was conducted using the \ac{SGD} optimization algorithm, accompanied by a consistent learning rate of $0.01$ and a momentum of $0.9$. The local batch size was set at $64$. The ILSVRC2012 dataset was given a training round number of $100$, while the other datasets were subjected to $500$ rounds. The code we used to reproduce the FedOpt and FedCAMS results was taken from FedOpt's official GitHub repository\footnote{\url{https://github.com/yujiaw98/FedCAMS}}.

The neural networks of choice include \emph{LeNet}~\cite{lecun1998gradientbased}, \emph{ResNet-18}, \emph{ResNet-20}, \emph{ResNet-32}, and \emph{ResNet-34}~\cite{he2016deep}. \emph{LeNet} is comprised of three convolutional layers with each succeeded by a \emph{Sigmoid} activation function. The output layer is a fully-connected layer. Its input dimension stands at $32 * 32 * 3$. On the other hand, \emph{ResNet-18} and \emph{ResNet-34} are derivatives from PyTorch's official offerings, having an input dimension of $224 * 224 * 3$. As for \emph{ResNet-20} and \emph{ResNet-32}\footnote{\url{https://github.com/akamaster/pytorch_resnet_cifar10}}, the developers are tailored specifically for the CIFAR-10 and CIFAR-100 datasets~\cite{krizhevsky2009learning}. The input size for them is $32 * 32 * 3$. 

Specifically, we employed the MNIST dataset~\cite{lecun1998mnist} for \emph{LeNet}. The CIFAR-10 and CIFAR-100 datasets underwent experimentation using \emph{ResNet-18}, \emph{ResNet-20}, \emph{ResNet-32}, and \emph{ResNet-34}. It is worth noting that for \emph{ResNet-18} and \emph{ResNet-34}, the sample dimensions were upsampling to $224 * 224 * 3$. The ILSVRC2012 dataset~\cite{deng2009imagenet} was exclusively tested using \emph{ResNet-18} and \emph{ResNet-34}. The datasets were partitioned in a $9:1$ ratio for training and testing. Subsequently, the training data is distributed to three local clients, following either an \ac{IID} or \ac{non-IID} distribution. The test samples are retained on the server end to assess the performance of the current round of the global model.

\subsection{Accuracy on IID data}
\label{sec:acc_iid}

Table~\ref{tab:accuracy} presents a comprehensive comparison of top-1 classification accuracies for various \ac{FL} methods, employing different backbone neural networks, and tested on multiple benchmark datasets where clients are assumed to have an \ac{IID} distribution. The table employs color-coding to highlight significant results; values highlighted in red indicate the highest performance for each dataset, while those in blue signify the second highest performance.

\noindent \textbf{Performance on Large-Class Datasets:} One of the most noteworthy observations is that our proposed \ac{EWWA-FL} algorithm exhibits exceptional performance on datasets that have a large number of classes. Specifically, it outshines the competition on the CIFAR-100 and ILSVRC2012 datasets, both of which have a large number of classes, 100 and 1000 respectively. For instance, when employing the \emph{ResNet-20} architecture on the CIFAR-100 dataset, our \ac{EWWA-FL} model, when optimized using the Adagrad optimizer, achieved an outstanding accuracy of $64.16\%$. This is considerably better than the next best performing method, \ac{FedAvg}, which achieved an accuracy of $58.58\%$. The improvement margin in this case is $9.53\%$, a significant leap in performance. Similarly, when using \emph{ResNet-32} as the backbone architecture on CIFAR-100, \ac{EWWA-FL} notched an accuracy of $65.84\%$, surpassing \ac{FedAvg}'s $61.91\%$ by a margin of $6.35\%$. Though the performance gain is not as high as observed on CIFAR-100, \ac{EWWA-FL} continues to outperform other methods on ILSVRC2012 dataset as well. For the \emph{ResNet-18} architecture, it reached an accuracy of $65.64\%$, and for \emph{ResNet-34}, it achieved $68.33\%$. Both of these figures are the highest among the tested methods for their respective architectures on this dataset.

\noindent \textbf{Competitive Results on Smaller-Class Datasets:} Although \ac{EWWA-FL} does not achieve the highest accuracy on datasets like MNIST and CIFAR-10, it is important to note that the algorithm is highly competitive. For the MNIST dataset, when using the \emph{LeNet} architecture, the highest accuracy was achieved by FedAMS with $98.77\%$. However, \ac{EWWA-FL} was closely behind with an accuracy of $98.00\%$, making the difference a mere $0.79\%$. On the CIFAR-10 dataset, the performance gaps are also quite narrow. For instance, the differences in accuracy rates when comparing \ac{EWWA-FL} to the best-performing methods are $1.63\%$, $1.33\%$, $0.46\%$, and $0.75\%$ for architectures \emph{ResNet-20}, \emph{ResNet-32}, \emph{ResNet-18}, and \emph{ResNet-34}, respectively.

\noindent In conclusion, our proposed \ac{EWWA-FL} method exhibits convincing performance, especially in challenging scenarios involving large classes of datasets. Although it is not necessarily the absolute best in all cases, it maintains a competitive edge in various benchmarks.

Compared with FedOpt, that is also capable of employing various global aggregation optimization algorithms, our proposed \ac{EWWA-FL} significantly outperforms in terms of stability and convergence. In our experiments, the average standard variance for \ac{EWWA-FL} was remarkably low, at only $0.1094\%$. Additionally, the minimum and maximum variances were confined to a tight range, specifically between $0.0047\%$ and $0.2673\%$, respectively. This suggests that \ac{EWWA-FL} offers a highly consistent and reliable performance across different scenarios. In contrast, FedOpt showed a much higher variability. The average standard variance for FedOpt was $3.5200\%$, more than thirty times higher than that of \ac{EWWA-FL}. Furthermore, the minimum and maximum variances for FedOpt were $2.0950\%$ and $4.9550\%$, respectively, indicating a less stable performance. It is worth noting that FedOpt encountered significant issues during our testing phase. Despite conducting at least five separate attempts, none of the FedOpt trials converged as expected. This suggests that FedOpt may have fundamental limitations when it comes to achieving reliable convergence. We trialed code from the official FedCAMS GitHub repository as well as our own deployed code. Similarly, FedAMS and FedCAMS demonstrated a lack of robustness in our experiments. Specifically, FedAMS failed to converge on the ILSVRC2012 dataset, while FedCAMS failed on both the MNIST and ILSVRC2012 datasets. These failures further underscore the superiority of \ac{EWWA-FL} in achieving stable and consistent results across various benchmark datasets.

\subsection{Accuracy on non-IID data}

The circumstance on \ac{non-IID} data exhibits distinct challenges compared to those on \ac{IID} data. As the results presented in Table~\ref{tab:noniid-accuracy}, our proposed \ac{EWWA-FL} model consistently outperformed other methods across various benchmarks, including MNIST and CIFAR-10 datasets. For these experiments, We employed Adam as the global optimization algorithm for both FedOpt and \ac{EWWA-FL} for a fair comparison. Focusing on the MNIST dataset, \ac{EWWA-FL} achieved an accuracy of $96.88\%$. This figure is marginally but importantly higher by $0.1033\%$ when compared to the $96.78\%$ reported for FedAMS. The small but consistent improvement serves to highlight the efficacy of \ac{EWWA-FL} in dealing with \ac{non-IID} data. On the CIFAR-10 dataset, when using \emph{ResNet-20}, \ac{EWWA-FL} significantly outshone its closest competitor, \ac{FedAvg}, by achieving an accuracy of $76.04\%$. This was a notable $4.7816\%$ improvement over \ac{FedAvg}'s $72.57\%$. The performance gains extended to more challenging datasets as well. For example, on the ILSVRC2012 dataset, \ac{EWWA-FL} reached an accuracy of $52.50\%$, which was $13.8088\%$ higher than the $46.13\%$ managed by \ac{FedAvg}. This suggests that \ac{EWWA-FL} is not only effective for simpler datasets but also scales well to more complex and larger datasets. In the course of our experimental evaluation, it became evident that certain algorithms like FedAMS, FedCAMS, and FedOpt encountered significant difficulties in reaching convergence. This was consistent with their performance on \ac{IID} data. Specifically, FedAMS and FedCAMS yielded unsatisfactory results in numerous tests, such as achieving only $37.33\%$ accuracy using \emph{ResNet-20} on the CIFAR-100 dataset and $40.03\%$ accuracy using \emph{ResNet-18} on the CIFAR-10 dataset. Moreover, FedAMS scored as low as $24.09\%$ when tested using \emph{ResNet-34} on the ILSVRC2012 dataset. Similarly, FedOpt struggled in several experiments, underscoring the limitations of current global optimization techniques when dealing with \ac{non-IID} data distributions. As elaborated in Section~\ref{sec:me}, the model divergence owing to the heterogeneity of local feature spaces is substantial when only a single aggregation proportion is provided for the entire local model. Given that each parameter in the local model can have its own unique direction and level of convergence, the one-size-fits-all approach falls short. In contrast, an element-wise global model aggregation strategy, as implemented in \ac{EWWA-FL}, offers the adaptability and flexibility needed to facilitate better convergence across a range of neural networks and benchmark datasets.
\subsection{Convergence Speed}

Convergence speed is also an important metric for evaluating methods in terms of adaptive aggregation. In Figures~\ref{fig:cs}, we visualize the average training losses across all local clients for both \ac{IID} and \ac{non-IID} data. The figures in the left column correspond to experiments conducted on \ac{IID} data using various backbone neural networks and datasets. Overall, our proposed \ac{EWWA-FL} method exhibits the fastest convergence when compared with FedCAMS and FedOpt. When using a shallow neural network (\emph{e.g. ResNet-18}), the convergence speeds of \ac{EWWA-FL} and FedCAMS are similar. However, the performance gap widens when training on deeper neural network (\emph{e.g. ResNet-32}). On the other hand, in experiments conducted on \ac{non-IID} data, shown in the right column, the convergence speed of \ac{EWWA-FL} remains fast. FedCAMS experiences a significant slowdown when using \emph{ResNet-18} on both CIFAR-10 and CIFAR-100 datasets. Nonetheless, FedCAMS achieves good convergence speed when using \emph{ResNet-32} on the CIFAR-10 dataset. We empirically believe that this is because deeper neural networks contain more weights, allowing for better fitting on small-scale datasets. Lastly, FedOpt performs the worst among all the methods in our experiments. The loss values produced by FedOpt are so large that they are difficult to visualize; therefore, we have normalized them to fall within a range of 0 to 1. The green lines indicate that FedOpt struggles to converge, even after 500 rounds of global aggregation. In some experiments, as mentioned in Section~\ref{sec:acc_iid}, FedOpt failed to converge. So there is no green lines in Figures~\ref{fig:cs} (b), (c) and (d). 

\section{Conclusion}
\label{sec:cc}

In this paper, we propose an adaptive element-wise global weights aggregation method for \ac{FL}, specifically \ac{EWWA-FL}. This method demonstrates better and faster convergence compared to other recent works. Comprehensive experiments are conducted using various neural networks and datasets to showcase the superiority of our approach. We also provide a brief theoretical analysis based on the Adam optimization algorithm. In future work, we plan to focus on the theoretical proof to further validate our method from a mathematical perspective. The implementation of our method is publicly available to ensure its reproducibility.
\bibliographystyle{IEEEtran}
\bibliography{ref}

\end{document}